\documentclass[conference]{IEEEtran}
\IEEEoverridecommandlockouts
\usepackage{cite}
\usepackage{amsmath,amssymb,amsfonts}
\usepackage{algorithm} 
\usepackage{algpseudocode}
\algrenewcommand{\algorithmiccomment}[1]{\hfill\textit{// #1}}
\usepackage{graphicx}
\usepackage[caption=false,font=footnotesize]{subfig}
\graphicspath{{figures/}{./}}
\usepackage{booktabs}
\usepackage{multirow}
\usepackage{placeins}
\usepackage{textcomp}
\usepackage{hyperref}
\hypersetup{hidelinks}
\usepackage{balance}
\usepackage{url}
\makeatletter
\g@addto@macro{\UrlBreaks}{\do\-\do\_\do\.}
\makeatother
\usepackage{xcolor}
\usepackage{microtype}
\usepackage{stfloats}
\def\BibTeX{{\rm B\kern-.05em{\sc i\kern-.025em b}\kern-.08em
    T\kern-.1667em\lower.7ex\hbox{E}\kern-.125emX}}
\begin{document}

\title{\texttt{IndicIPR-QA:} A Benchmark for Extractive Question Answering on Indian IPR Laws}

\author{
\IEEEauthorblockN{M. Mohaiminul Islam}
\IEEEauthorblockA{
\textit{Department of Computer Science and Engineering} \\
\textit{South Asian University, New Delhi, India} \\
niloy@students.sau.ac.in
}
\and
\IEEEauthorblockN{Muhammad Abulaish\textsuperscript{*}, \textit{SMIEEE}\thanks{\textsuperscript{*}Corresponding author: abulaish@ieee.org}}
\IEEEauthorblockA{
\textit{Department of Computer Science and Engineering} \\
\textit{South Asian University, New Delhi, India} \\
abulaish@ieee.org
}
}

\maketitle

\begin{abstract}

Legal Question Answering (QA) over statutory texts is a challenging task in Natural Language Processing (NLP), especially in low-resource settings with complex legal systems. It requires answers that are precise and grounded in the law. However, to the best of current knowledge, no existing extractive QA benchmark provides exact answer spans from statutory texts across all seven Indian Intellectual Property Rights (IPR) Acts. This paper presents IndicIPR-QA\footnote{Dataset: \url{https://huggingface.co/datasets/niloycste68/IndicIPRExtractiveQA};
code: \url{https://github.com/niloycste/Indic-IPR-Dataset-}}, an extractive QA benchmark for Indian IPR laws. The proposed dataset contains 4,462 QA pairs from 1,116 statutory contexts across seven Acts. Questions and answers are generated using GPT-4o and validated through nine automatic quality checks, single and multi-model LLM evaluations, and a 200-pair human audit. Section-level partitioning prevents source-context overlap across the training, validation, and test sets. Further analysis examines answer span length, lexical overlap, and multi-clause structure across the corpus and individual Acts. BM25 achieves a token-level F1 score of $20.21$, while fine-tuned InLegalBERT reaches $85.02$ under a unified sliding-window inference protocol. The large performance gap between BM25 (20.21 F1) and InLegalBERT (85.02 F1) shows that simple word matching is not enough to answer these legal questions. Performance also varies across the seven Acts, indicating substantial differences in answer-extraction performance.
\end{abstract}


\section{Introduction}
\label{sec:intro}

Intellectual Property Rights (IPR) law covers many economically and socially important areas  including pharmaceutical innovation, creative works, agricultural plant varieties, and region specific products. In India, IPR law is covered by seven principal statutes: the Patents Act, 1970; the Trade Marks Act, 1999; the Copyright Act, 1957; the Designs Act, 2000; the Geographical Indications of Goods (Registration and Protection) Act, 1999; the Semiconductor Integrated Circuits Layout-Design Act, 2000; and the Protection of Plant Varieties and Farmers' Rights Act, 2001. Although these Acts belong to the same legal domain, each covers different subject matter and uses its own terminology, procedures, exceptions, and cross references. As a result, finding the relevant legal provision can be difficult for legal professionals, researchers, and other stakeholders. A startup may need to determine trademark registration requirements; an inventor may need the legal conditions for patentability; a farmer or breeder may need information on plant-variety rights; and a designer or copyright owner may need to identify available forms of protection or remedies. The required information is present in the legislation, but it is often buried in long multi-clause provisions, nested subsections, and specialised legal language. The challenge is therefore not simply accessing the law, but locating a precise and traceable answer within it.

Statutory Question Answering (QA) also creates a specific evaluation challenge. A plausible paraphrase may be unsuitable when the exact wording of the law is required, while returning an entire provision may be correct but unnecessarily long. A useful benchmark should therefore evaluate both whether the relevant provision is found and whether the answer span is selected correctly. These considerations motivate IndicIPR-QA's focus on answer in context validity, span minimality, source-separated splits, and item-level error analysis.

Extractive QA provides a suitable setting for this problem because each predicted answer can be traced directly to its source provision. However, existing QA benchmarks are mainly general domain, contract based, non extractive, or focused on court judgments rather than Indian statutory IPR law. IndicIPR-QA addresses this gap by providing a benchmark for developing and evaluating statutory search and grounded legal QA systems. It is intended as a research benchmark, not as a deployed legal advice system. The major contributions of this paper can be outlined as follows:
\begin{enumerate}
    \item IndicIPR-QA contains 4,462 extractive QA pairs from 1,116 statutory contexts covering seven major Indian IPR Acts. The quality of the dataset is validated through nine automatic checks, LLM based evaluation, and a 200 pair human audit by two law graduates.
    
    
    \item The analysis studies answer length, word overlap between questions and legal texts, multi-clause answers, and differences in answer extraction performance across the seven Acts.
    
    \item Four BERT-based encoders and three open LLMs are benchmarked. BM25 achieves a token level F1 of 20.21, while InLegalBERT reaches 85.02, giving a 64.81 point gap that shows that simple word matching is not enough for this task (Section~\ref{sec:baselines}).
\end{enumerate}

\section{Related Work}
\label{sec:related}

Existing research has introduced several benchmark datasets for question answering and reasoning over legal texts, covering tasks such as extractive QA, statutory retrieval, legal reasoning, contract understanding, and judgment analysis. SQuAD established the standard extractive QA setting, where the answer is selected as a span from a given passage~\cite{rajpurkarSQuAD100000Questions2016}, and SQuAD 2.0 later added unanswerable questions~\cite{rajpurkar2018know}. Although these datasets have had a major impact, they use general-domain text and do not capture the structure, terminology, and constraints of legal statutes. Legal QA datasets cover several other settings. CUAD provides expert annotated answer spans from commercial contracts~\cite{hendrycksCUADExpertAnnotatedNLP2021}, while JEC-QA contains multiple choice questions from the Chinese judicial examination~\cite{zhongJECQALegalDomainQuestion2020}. PrivacyQA, LLeQA, and VLQA focus on privacy policies, legal consultation, and questions from Vietnamese citizens, respectively~\cite{ravichander-etal-2019-question,louis2024interpretable,nguyen2025vlqa}. Statutory retrieval benchmarks such as COLIEE~\cite{rabelo2022overview} and BSARD~\cite{louis2022statutory} focus on finding relevant legal provisions, while IPBench evaluates LLM knowledge across a broader bilingual set of intellectual property tasks~\cite{wang2025ipbench}.

These datasets also differ in how they evaluate legal QA. Retrieval benchmarks mainly measure whether the relevant legal provision is found, whereas extractive QA also requires the correct start and end positions of the answer within that provision. Multiple choice and generative tasks do not directly measure whether the predicted answer matches the source text. Extractive QA is therefore useful when answers need to remain directly grounded in the law. In this setting, answer quality depends not only on finding the correct provision but also on selecting an appropriate answer span.

Research on Indian legal NLP has explored domain-specific language modelling and tasks such as legal statute identification, judgment segmentation, and court appeal prediction~\cite{paulPretrainedLanguageModels2023}. Statutes and judgments have different structures and purposes: statutes define rules through sections, definitions, provisions, and cross-references, while judgments explain how those rules are applied to specific disputes. LLM assisted data construction provides a scalable alternative to fully manual annotation, but it can also introduce incorrect spans, repeated question patterns, and inconsistent answer boundaries. These issues make careful quality control important when building automatically generated QA datasets.

Overall, existing resources cover legal retrieval, reasoning, contracts, judgments, and broader intellectual property tasks. To our knowledge, existing resources do not provide an extractive QA benchmark covering all seven major Indian IPR Acts. IndicIPR-QA addresses this gap by providing extractive QA pairs with answer spans directly grounded in Indian statutory text, along with quality checks.

\section{Dataset Construction}
\label{sec:dataset}

\begin{figure*}[!t]
\centering
\includegraphics[height=7.35cm, width=16cm]{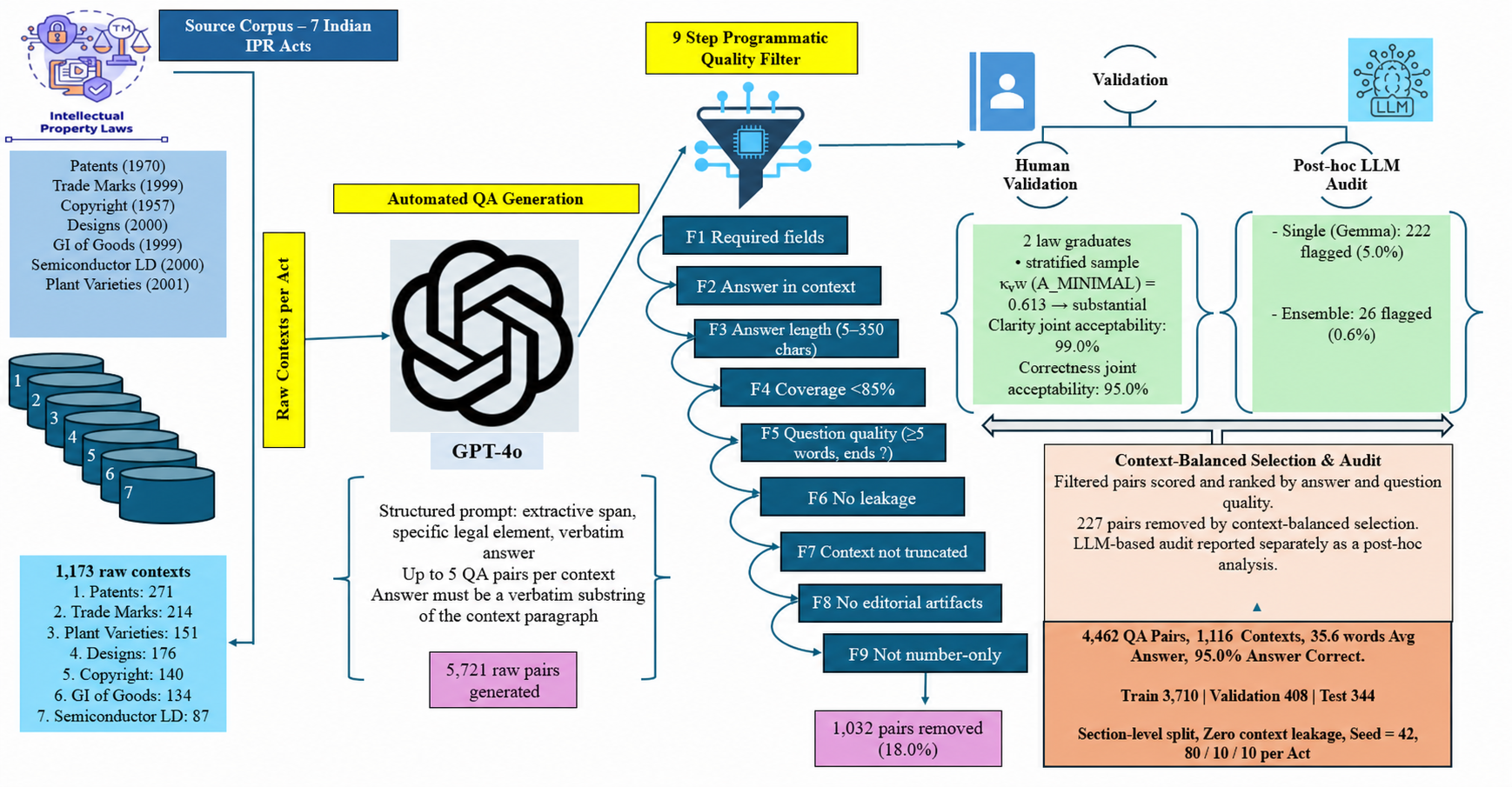}
\caption{End-to-end IndicIPR-QA construction. Human-audit percentages denote
joint acceptability on the 200-pair sample; LLM scores are post-hoc diagnostics.}
\label{fig:pipeline}
\end{figure*}

The dataset construction proceeded in five phases:
corpus creation, QA generation, quality control, section level splitting, and audit/validation. Fig.~\ref{fig:pipeline} and Algorithm~\ref{alg:dataset_construction} provide an overview of the full pipeline, from raw statute text to the curated benchmark.

\begin{algorithm}[t]
\caption{IndicIPR-QA: From raw statutes to curated benchmark}
\label{alg:dataset_construction}
\scriptsize
\begin{algorithmic}[1]

\Statex \textbf{Input:} Act PDF set $\mathcal{P}=\{P_1,\ldots,P_7\}$; chunk size $\delta=1400$
\Statex \textbf{Output:} Curated dataset $\mathcal{D}$, retained contexts $\mathcal{C}^{*}$, and splits $(\mathcal{D}_{\mathrm{tr}},\mathcal{D}_{\mathrm{va}},\mathcal{D}_{\mathrm{te}})$

\Statex \textit{\textbf{Step 1: Corpus extraction}}
\State $\mathcal{C} \gets \emptyset$
\ForAll{$P_i \in \mathcal{P}$}
    \State $T_i \gets \Call{ExtractText}{P_i}$ \Comment{TOC removal, body start detection}
    \State $T_i \gets \Call{CleanText}{T_i}$ \Comment{Remove editorial lines, normalize whitespace}
    \State $\mathcal{C} \gets \mathcal{C} \cup \Call{ChunkContexts}{T_i,\delta}$ \Comment{Sentence-safe chunking}
\EndFor
\State $\mathcal{C} \gets \Call{ValidateContexts}{\mathcal{C}}$ \Comment{$1{,}173$ valid contexts}

\Statex \textit{\textbf{Step 2: QA pair generation}}
\State $\mathcal{R} \gets \emptyset$
\ForAll{$c \in \mathcal{C}$}
    \State $\mathcal{R} \gets \mathcal{R} \cup \Call{GenerateQA}{c}$ \Comment{GPT-4o; up to 5 extractive QA pairs}
\EndFor
\Statex \quad\textit{// $5{,}721$ raw candidate pairs}

\Statex \textit{\textbf{Step 3: Quality control and context-balanced selection}}
\State $\mathcal{R} \gets \Call{ApplyFilters}{\mathcal{R},F_1,\ldots,F_9}$ \Comment{$4{,}689$ pairs after automatic filtering}
\State $\mathcal{D} \gets \Call{ContextBalancedSelect}{\mathcal{R},4462}$ \Comment{Score-based selection}
\State $\mathcal{C}^{*} \gets \Call{RetainCoveredContexts}{\mathcal{C},\mathcal{D}}$ \Comment{$1{,}116$ retained statutory contexts}

\Statex \textit{\textbf{Step 4: Section-level dataset splitting}}
\State $\mathcal{G} \gets \Call{GroupBySection}{\mathcal{D}}$ \Comment{$681$ unique statute sections}
\State $(\mathcal{D}_{\mathrm{tr}},\mathcal{D}_{\mathrm{va}},\mathcal{D}_{\mathrm{te}})
\gets \Call{StratifiedSplit}{\mathcal{G}}$ \Comment{Per-Act split, seed 42; 3,710/408/344}

\Statex \textit{\textbf{Step 5: Audit and validation}}
\State \Call{LLMJudgeAudit}{$\mathcal{D},\textsc{Gemma-3:12B}$} \Comment{Single-judge audit}
\State \Call{LLMJudgeAudit}{$\mathcal{D},\textsc{Ensemble-3}$} \Comment{Three-model ensemble}
\State \Call{HumanValidation}{$\mathcal{D}_{\mathrm{te}}$} \Comment{200-pair sample, 2 annotators}

\State \Return $\mathcal{D},\mathcal{C}^{*},(\mathcal{D}_{\mathrm{tr}},\mathcal{D}_{\mathrm{va}},\mathcal{D}_{\mathrm{te}})$

\end{algorithmic}
\end{algorithm}

\subsection{Source Corpus and QA Generation}
The full text of all seven Indian IPR Acts is collected from official Government of India PDFs available on the India Code portal (\url{https://www.indiacode.nic.in/}) using pdfplumber. During preprocessing, table of contents pages were removed, editorial amendment markers such as ``Ins.,'' ``Subs.,'' and ``w.e.f.'' were deleted, and whitespace was normalised. The cleaned text is then divided into 1,173 paragraph-level contexts, each about 1,400 characters long, while preserving sentence and section boundaries. These chunks are called statutory contexts throughout the paper.

For each context, GPT-4o~\cite{hurstGPT4oSystemCard2024} is prompted to generate up to five QA pairs. The prompt requires a structured JSON output, questions based on specific facts or legal provisions, and answers copied exactly from the context. It also prohibits paraphrased answers, yes/no or vague questions, answer leakage, and near-duplicate questions. The full GPT-4o generation prompt and generation script are provided in the public repository.\footnote{\url{https://github.com/niloycste/Indic-IPR-Dataset-}} This process generated 5,721 raw QA candidates.

\subsection{Nine-Step Quality Control Pipeline}
\label{sec:pipeline}

All 5,721 raw pairs pass through nine sequential filters. F1 checks for non empty fields; F2 verifies that the answer is an exact span from the context; F3 keeps answers from 5 to 350 characters; F4 keeps answers covering less than 85\% of the context; F5 requires questions to contain at least five words, end with a question mark, and avoid vague section-summary phrasing; F6 removes questions that contain the answer; F7 removes incomplete contexts; F8 removes editorial amendment markers from answers; and F9 removes standalone 1 to 4 digit answers. Together, these filters remove schema errors, unsupported answers, overly short or overly long spans, weak questions, and extraction artefacts.

F3 and F4 remove 595 and 417 pairs, respectively, accounting for 1,012 of the 1,032 automatic exclusions. This shows that answer span length is the main source of automatic filtering. The remaining 4,689 pairs then undergo context balanced selection. Candidates are ranked based on the length of questions and answers while maintaining coverage across contexts. This produces the final 4,462 QA pairs from 1,116 statutory contexts. Removing this selection step reduces test F1 by 1.34 points (Table~\ref{tab:model_ablation}). Finally, section level partitioning creates training, validation, and test sets of 3,710/408/344 QA pairs, with no shared source context across the splits.

\subsection{Pipeline Ablation Study}

To assess the contribution of each filter, a leave one filter out ablation is performed while keeping the remaining eight filters active. Table~\ref{tab:ablation} reports the number of additional candidates admitted when each filter is disabled.

The results show that F3 and F4 have the largest effect on the accepted set. Disabling F3 admits 368 additional potentially noisy pairs (+7.8\% relative to the 4,689 pair baseline), while disabling F4 admits 417 additional pairs (+8.9\%). The main automatically detected issue is answer span length: generated spans are often either too short to be informative or too long, covering most of the source context.

\subsection{Model Level Filter Ablation}
\label{sec:model_ablation}

To test whether the quality control pipeline affects model performance, InLegalBERT is trained on six different training-data variants while keeping the validation and test sets fixed. All variants in Table~\ref{tab:model_ablation}, including the clean-selected reference, are trained separately in this ablation study. The main baseline model in Table~\ref{tab:all_baselines} is trained separately.

Table~\ref{tab:model_ablation} shows that the largest drop in F1 comes from the selection step rather than from any single filter. Removing context balanced selection (heuristic all) lowers F1 by 1.34 points. Using all structurally valid raw candidates lowers F1 by 0.67 points, while removing F4 alone lowers it by 0.21 points.

The other results are less clear. Removing the answer length filter F3 alone increases test F1 slightly by 0.45 points, and removing F3 and F4 together increases it by 0.47 points. Since these results come from single seed runs and no statistical significance test was performed, the ablation results are treated as diagnostic rather than as proof that each filter improves model performance. The filters are retained because they enforce the answer validity and span quality constraints captured by the automatic checks in Table~\ref{tab:ablation}. Model F1 alone is not enough to measure these properties.

\begin{table}[!t]
\centering
\caption{Quality-filter analysis of the 5,721 raw GPT-4o generated candidate
QA pairs. Sequential filtering applies filters F1 to F9 in order. Per-filter
ablation omits each filter individually while keeping the remaining
filters active.}
\label{tab:ablation}
\scriptsize
\setlength{\tabcolsep}{2.2pt}
\resizebox{\columnwidth}{!}{%
\begin{tabular}{@{}llrrrr@{}}
\toprule
\multirow{2}{*}{\textbf{ID}} &
\multirow{2}{*}{\textbf{Filter criterion}} &
\multicolumn{2}{c}{\textbf{Sequential filtering}} &
\multicolumn{2}{c}{\textbf{Per-filter ablation}} \\
\cmidrule(lr){3-4}\cmidrule(l){5-6}
& & \textbf{Removed} & \textbf{Retained}
& \textbf{Recovered pairs} & \textbf{Accepted} \\
\midrule
F1 & Required fields               &   0   & 5,721 &   0 (0.0\%) & 4,689 \\
F2 & Answer in context             &   0   & 5,721 &   0 (0.0\%) & 4,689 \\
F3 & Answer length 5--350$^\dagger$ 
   & 595 & 5,126 & 368 (7.8\%) & 5,057 \\
F4 & Coverage $<$85\%$^\ddagger$    
   & 417 & 4,709 & 417 (8.9\%) & 5,106 \\
F5 & Question quality              &  13   & 4,696 &  13 (0.3\%) & 4,702 \\
F6 & No answer leakage             &   1   & 4,695 &   1 (0.0\%) & 4,690 \\
F7 & Context not truncated         &   3   & 4,692 &   3 (0.1\%) & 4,692 \\
F8 & No editorial artefacts        &   3   & 4,689 &   3 (0.1\%) & 4,692 \\
F9 & Answer not num-only           &   0   & 4,689 &   0 (0.0\%) & 4,689 \\
\midrule
\multicolumn{2}{@{}l}{}
& 1,032 & 4,689 & \multicolumn{2}{c}{} \\
\bottomrule
\end{tabular}%
}
\begin{flushleft}
{\footnotesize
$^\dagger$F3 removes 595 pairs during sequential filtering. When F3 is omitted,
only 368 pairs are recovered because the remaining 227 F3-rejected candidates
are still removed by one or more of the other active filters.\\
$^\ddagger$F4 removes 417 pairs during sequential filtering. When F4 is omitted,
all 417 pairs are recovered, increasing the accepted set from 4,689 to 5,106.
}
\end{flushleft}
\end{table}

\begin{table}[!t]
\centering
\caption{Model level filter ablation using InLegalBERT on different training data variants. All archived checkpoints are evaluated with the unified decoder; $\Delta$F1 is relative to the clean selected run.}
\label{tab:model_ablation}

\scriptsize
\setlength{\tabcolsep}{10.5pt}
\renewcommand{\arraystretch}{0.92}

\begin{tabular}{@{}lccc@{}}
\toprule

\textbf{Training variant} 
& \textbf{F1} 
& \textbf{Char-F1} 
& \textbf{$\Delta$F1} \\

\midrule

clean selected (ablation control)        & 84.11 & 86.30 & ---     \\
without F3+F4 combined       & 84.58 & 87.25 & $+$0.47 \\
without F3 (answer length)   & 84.56 & 87.13 & $+$0.45 \\
without F4 (coverage)        & 83.90 & 86.66 & $-$0.21 \\
raw candidates (unfiltered)  & 83.44 & 85.64 & $-$0.67 \\
heuristic all (no selection) & 82.77 & 86.59 & $-$1.34 \\

\bottomrule
\end{tabular}
\end{table} 

\subsection{LLM as Judge Quality Verification}
\label{sec:llm_judge}

To obtain an additional automated assessment of the 4,462 retained QA pairs beyond rule based filtering, an LLM as judge protocol is applied under two different configurations, and their outputs are compared. These scores are used only as auxiliary diagnostic indicators and not as a replacement for human evaluation.

Gemma~3:12B~\cite{gemmateamGemma3Technical2025} evaluates four criteria: Q\_SPECIFIC (the question targets a specific legal element), A\_SUFFICIENT (the answer fully addresses the question), A\_MINIMAL (the answer is the shortest correct span), and LEGAL\_RELEVANT (the question concerns substantive legal content). Scores are 1 (poor), 2 (acceptable), or 3 (good), and a pair is flagged if any criterion receives a score of 1.

Gemma~3:12B, Llama~3.1:8B~\cite{teamLlama3Herd2024}, and
Mistral~7B~\cite{jiangMistral7B2023} evaluate the same four
criteria independently. When at least two models assign the
same score, that score is used as the ensemble score. The
released code contains the complete aggregation rule.
Table~\ref{tab:llm_judge_compare} compares the evaluation
results of the single judge and the three-model ensemble.

\begin{table}[!t]
\centering
\caption{LLM as Judge evaluation with a single judge and a three model ensemble. Scores range from 1 (poor/flagged) to 3 (good).}
\label{tab:llm_judge_compare}

\scriptsize
\setlength{\tabcolsep}{6pt}
\renewcommand{\arraystretch}{0.92}

\begin{tabular}{@{}lcccc@{}}
\toprule

& \multicolumn{2}{c}{\textbf{Single Judge}} 
& \multicolumn{2}{c}{\textbf{Ensemble}} \\

\cmidrule(lr){2-3}
\cmidrule(l){4-5}

\textbf{Criterion} 
& \textbf{Avg. score} 
& \textbf{Flagged} 
& \textbf{Avg. score} 
& \textbf{Flagged} \\

\midrule

Question specificity & 2.92 & 1.1\% & 2.93 & 0.0\% \\
Answer sufficiency   & 2.92 & 2.8\% & 2.99 & 0.4\% \\
Answer minimality    & 2.52 & 4.2\% & 2.85 & 0.4\% \\
Legal relevance      & 2.97 & 0.7\% & 3.00 & 0.0\% \\

\midrule

Flagged QA pairs 
& \multicolumn{2}{c}{222 pairs (5.0\%)} 
& \multicolumn{2}{c}{26 pairs (0.6\%)} \\

\bottomrule
\end{tabular}

\end{table}

The largest difference is in answer minimality. The single judge flags more long answer spans, while the ensemble is more permissive. The 26 pairs flagged by the ensemble include incomplete spans and long section level answers. The three judges give the same score on all four criteria for only 8.0\% of the pairs, showing that their judgments can differ. Therefore, the LLM as Judge results are treated as diagnostic evidence rather than as final measures of dataset quality.

\subsection{Human Validation Study}
\label{sec:human_val}

Two law graduates independently audited 200 test pairs across all seven Acts (seed~42: copyright 25, designs 24, GI 32, patents 29, plant varieties 28, semiconductor 30, and trade marks 32). A sample of 200 pairs is selected as a practical post hoc audit size while maintaining coverage across all seven Acts. The audit is diagnostic rather than a precise estimate of quality across all 4,462 pairs or of expert IPR annotation accuracy. Pairs were rated from 1 (poor) to 3 (good) for question clarity, answer correctness, and answer minimality; agreement is measured using unweighted and quadratic-weighted Cohen's $\kappa$~\cite{cohenCoefficientAgreementNominal1960,cohen1968weighted}.

Table~\ref{tab:kappa} reports human inter-annotator agreement and joint acceptability. Joint acceptability is 99.0\% for clarity and 95.0\% for correctness. Most correctness disagreements are between scores 3 and 2 ($\kappa_\text{QW}=0.328$). Minimality shows higher agreement ($\kappa_\text{QW}=0.613$), with 75.5\% joint acceptability.

\FloatBarrier

\begin{table}[!t]
\centering
\caption{Human inter-annotator agreement on 200 QA pairs.
$\kappa_\text{UW}$ = unweighted Cohen's $\kappa$; $\kappa_\text{QW}$ =
quadratic-weighted Cohen's $\kappa$.}
\label{tab:kappa}

\scriptsize
\setlength{\tabcolsep}{6pt}
\renewcommand{\arraystretch}{0.92}

\begin{tabular}{@{}lccc@{}}
\toprule

\textbf{Criterion}
& \textbf{$\kappa_\text{UW}$}
& \textbf{$\kappa_\text{QW}$}
& \textbf{Accept.\ \%} \\

\midrule

Question Clarity
& $-0.008$
& 0.051
& 99.0\%$^\dagger$ \\

Answer Correctness
& 0.117
& 0.328
& 95.0\% \\

Answer Minimality
& 0.312
& 0.613
& 75.5\%$^\ddagger$ \\

\bottomrule
\end{tabular}

\vspace{2pt}

\begin{flushleft}
\footnotesize
$^\dagger$ The near-zero $\kappa$ for Q\_CLEAR reflects a ceiling effect: both annotators gave scores $\geq 2$ to 198 of 200 pairs (99.0\%).

\vspace{2pt}

$^\ddagger$ Joint acceptability requires both annotators to give a score $\geq 2$. For minimality, 75.5\% of pairs meet this threshold, while 82.0\% are acceptable to at least one annotator. Both annotators rated 36 pairs as non-minimal (score 1). These pairs remain in the released dataset because the audit is conducted after dataset construction and is not used for filtering.
\end{flushleft}

\end{table}

\begin{figure}[!t]
\centering
\includegraphics[width=0.98\linewidth]{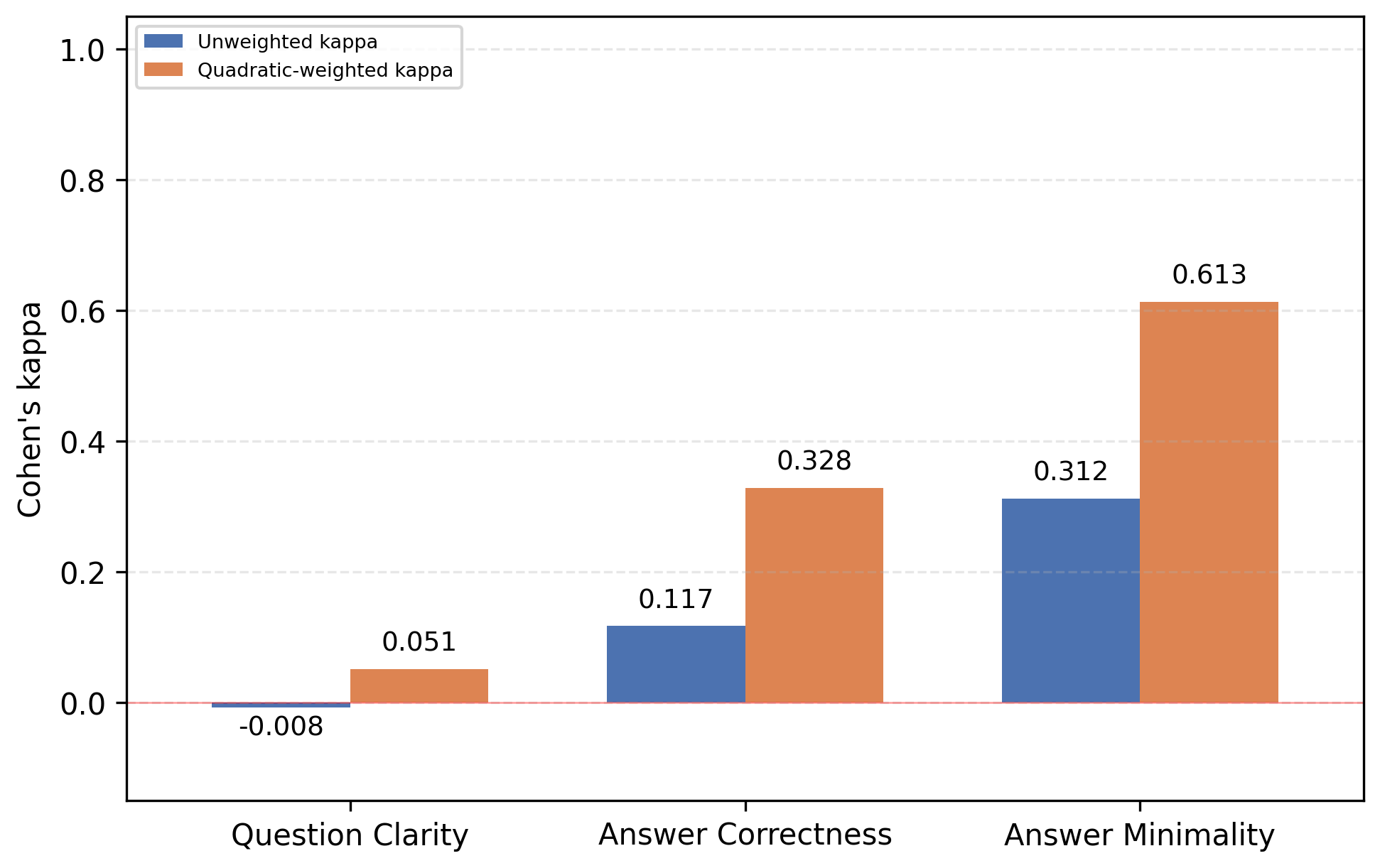}
\caption{Criterion-wise inter-annotator agreement for the human validation
study. Answer minimality has the strongest agreement, while question clarity
shows a ceiling effect due to skewed marginal distributions.}
\label{fig:kappa_plot}
\end{figure}

This highlights the difficulty of selecting the shortest correct span for
long, multi-clause answers. Since the audit interface truncated long
contexts, these results should be viewed as diagnostic rather than as a
complete measure of dataset quality. Figure~\ref{fig:kappa_plot}
summarises agreement across the three criteria.

\FloatBarrier

\section{Dataset Analysis}
\label{sec:analysis}

This section examines answer length, lexical overlap, span structure, and model performance across the dataset and the seven Acts.

\subsection{Corpus Level Statistics}

Table~\ref{tab:stats_comparison} compares IndicIPR-QA with established benchmarks. Its average answer length is 35.6 words, about 11$\times$ SQuAD's 3.2 words and 19\% longer than CUAD's 30.0 words, which may increase span boundary complexity. The dataset is fully extractive, contains no unanswerable questions, and uses section level train test separation.

\begin{table*}[!t]
\centering
\caption{Comparison of IndicIPR-QA with related QA benchmarks
(SQ1 = SQuAD 1.1~\cite{rajpurkarSQuAD100000Questions2016}; SQ2 = SQuAD 2.0~\cite{rajpurkar2018know}; CUAD~\cite{hendrycksCUADExpertAnnotatedNLP2021}; JEC = JEC-QA~\cite{zhongJECQALegalDomainQuestion2020}; IPB = IPBench~\cite{wang2025ipbench}).
``---'' indicates unavailable or not directly comparable statistics. Values follow the reporting conventions of each benchmark's source and may use different dataset split definitions.}
\label{tab:stats_comparison}
\setlength{\tabcolsep}{2pt}
\begin{tabular}{@{}lcccccc@{}}
\toprule
\textbf{Property} & \textbf{SQ1} & \textbf{SQ2} &
\textbf{CUAD} & \textbf{JEC} &
\textbf{IPB} & \textbf{IndicIPR-QA (Ours)} \\
\midrule
QA pairs        & 87.6k & 150k & 13.1k & 26.4k & 10.4k & \textbf{4.5k} \\
Contexts        & 18.9k & 35k  & 510   & ---   & ---   & \textbf{1,116} \\
QA/context      & 4.6   & 4.3  & 25.7  & ---   & ---   & \textbf{4.0} \\
Ans len (wds)   & 3.2   & 3.2  & 30.0  & MCQ   & Mixed & \textbf{35.6} \\
Ctx len (wds)   & 117   & 117  & 200   & ---   & 200   & \textbf{171.3} \\
Q len (wds)     & 10.1  & 10.1 & 9.5   & ---   & ---   & \textbf{11.9} \\
Human review    & Yes   & Yes  & Yes   & Yes   & Yes   & \textbf{Yes}$^\dagger$ \\
Extractive      & Yes   & Yes  & Yes   & No    & No    & \textbf{Yes} \\
Unanswerable    & No    & Yes  & Yes   & No    & No    & \textbf{No} \\
Domain          & General & General & Contracts & Chinese Bar Exam & US/Chinese IP Law & \textbf{Indian IPR} \\
\bottomrule
\end{tabular}
\begin{flushleft}
{\footnotesize 
$^\dagger$Audited on 200 test pairs by two law graduates; see Section~\ref{sec:human_val}.
}
\end{flushleft}
\end{table*}

\subsection{Per-Act Distribution and Performance}

Table~\ref{tab:per_act} reports per-Act statistics and InLegalBERT test performance. Performance varies substantially across the seven Acts.

\begin{table}[!t]
\centering
\caption{Per-Act IndicIPR-QA statistics, sorted by test F1. $n_{\mathrm{all}}$ and $n_{\mathrm{test}}$ denote full-dataset and test-set counts; Q-Ctx denotes stopword removed question-context unigram overlap.}
\label{tab:per_act}
\scriptsize
\begin{tabular}{@{}lrrrrr@{}}
\toprule
\textbf{Act} & $\mathbf{n_{all}}$ & $\mathbf{n_{test}}$ & \textbf{Avg. ans. (words)} &
\textbf{Q-Ctx} & \textbf{F1} \\
\midrule
Designs      & 173   & 24 & 43.4 & 0.320 & 100.00 \\
Patents      & 1,143 & 69 & 35.8 & 0.429 & 98.40 \\
Semicond.    & 585   & 47 & 22.1 & 0.418 & 91.53 \\
Trade Marks  & 1,106 & 88 & 31.9 & 0.416 & 83.83 \\
GI           & 551   & 51 & 51.8 & 0.458 & 83.66 \\
Copyright    & 349   & 25 & 62.3 & 0.447 & 79.41 \\
Plant Var.   & 555   & 40 & 21.9 & 0.636 & 53.20 \\
\bottomrule
\end{tabular}
\end{table}

Test F1 ranges from 100.00 for Designs to 53.20 for Plant Varieties. The perfect score for Designs should be viewed with caution because the test set contains only 24 pairs. Across the seven Acts (n = 7), question-context unigram overlap shows a strong negative correlation with F1 ($r = -0.928$). Given the small number of Acts, this correlation should be interpreted as exploratory rather than as evidence of a general relationship. Plant Varieties have the highest overlap (0.636) but the lowest F1. One possible reason is the presence of specialised biological and agricultural terms, although this is not directly tested. Copyright has more multi-clause answers (53.9\%) than Patents (13.7\%) and also has lower F1 (79.41 versus 98.40).

The answers cover several legal topics, including authority (29.4\%), procedure (12.0\%), rights (9.5\%), conditions (7.9\%), definitions (5.8\%), duration (2.3\%) and penalties (1.7\%), while 31.4\% fall into other categories. The average question to context word overlap is 0.448 after removing common stopwords. Also, 19.9\% of answers contain semicolons and 5.9\% contain ``subject to,'' showing that many answers contain multiple clauses. When the answers are divided into short and long groups using the median length of 33 words, InLegalBERT F1 decreases from 86.24 for short answers to 83.78 for long answers. Multi-clause answers also have lower boundary-sensitive EM than single-clause answers (56.10 versus 66.03). Answers that appear later in the context have slightly lower F1 than earlier answers (84.21 versus 85.83). Overall, answer length, the number of clauses, and the answer's position in the context are associated with extraction performance.

\FloatBarrier

\section{Baseline Experiments \& Results}
\label{sec:baselines}

Four BERT based encoder models are fine tuned on the IndicIPR-QA training split (3,710 QA pairs) using the BertForQuestionAnswering architecture~\cite{devlin2019bert}. This architecture predicts the start and end positions of the answer span in the input sequence: [CLS] question [SEP] context [SEP]. The models cover different levels of domain specialisation: BERT base uncased~\cite{devlin2019bert} (general English), LegalBERT-US~\cite{chalkidisLEGALBERTMuppetsStraight2020} (US legal text), InCaseLawBERT~\cite{paul2022pre} (Indian court judgments), and InLegalBERT~\cite{paulPretrainedLanguageModels2023} (Indian legal text).

All encoder models use the same settings: AdamW optimiser, learning rate $3\times10^{-5}$ with 50 warm-up steps and linear decay, weight decay 0.01, effective batch size 32, maximum sequence length 384, document stride 128, 3 training epochs, and random seed 42. The same settings are used for all encoder models to ensure a controlled comparison. The best checkpoint is selected using validation F1. During inference, long contexts are divided into overlapping windows using a stride of 128. A constrained decoder then selects the highest-scoring valid answer span across all windows and returns the exact text from the context. All encoder results reported here are recomputed from the archived checkpoints using this same inference procedure; no model is retrained.

Three open instruction tuned LLMs are also evaluated in zero shot and 3 shot settings: Llama~3.1:8B~\cite{teamLlama3Herd2024}, Mistral~7B~\cite{jiangMistral7B2023}, and Qwen2.5:7B~\cite{qwenQwen25TechnicalReport2024}. These models come from different model families and have similar sizes of about 7 to 8B parameters. In the zero shot setting, each model receives the statute passage and question and is instructed to copy the answer exactly from the context. In the 3 shot setting, three example QA pairs from the training set are provided as demonstrations.

As a simple lexical baseline, BM25~\cite{robertsonProbabilisticRelevanceFramework2009} is also evaluated. For a document $D$ and query $Q$, $f(q,D)$ denotes the frequency of token $q$, $|D|$ the document length, and $\overline{|D|}$ the average document length. The standard values $k_1 = 1.5$ and $b = 0.75$ are used. In the oracle window setting, BM25 is given the correct answer containing context, isolating within context retrieval. The relevance score is
\begin{equation}
\mathrm{BM25}(D,Q) =
\sum_{q \in Q} \mathrm{IDF}(q)
\frac{f(q,D)(k_1+1)}
{f(q,D)+k_1\!\left(1-b+b\tfrac{|D|}{\overline{|D|}}\right)}
\end{equation}

\subsection{Results}
\label{sec:results}

\begin{figure}[!htbp]
\centering
\includegraphics[width=\linewidth]{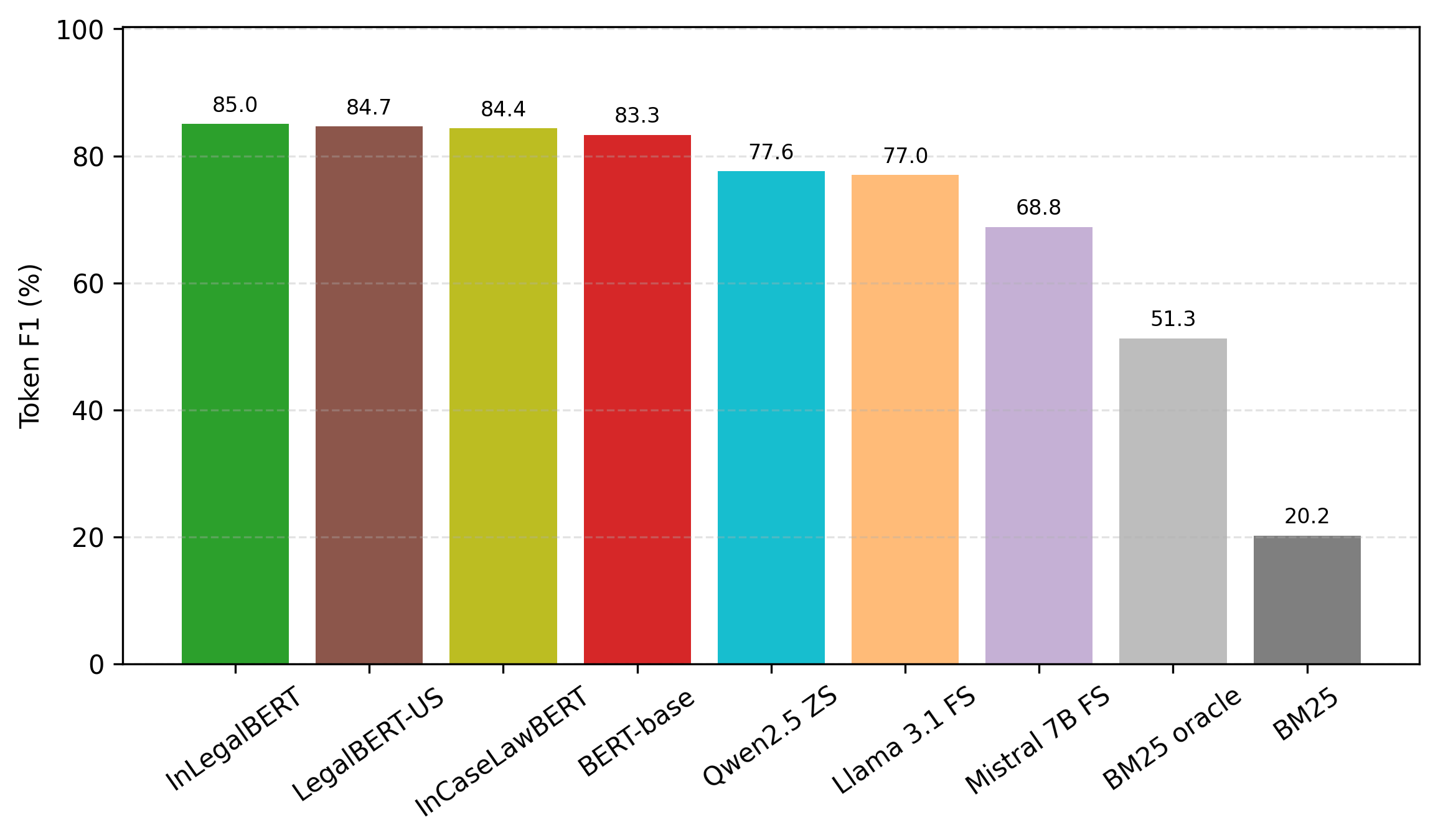}
\caption{Token-level F1 across encoder and lexical baselines and the best
prompted setting for each open LLM on the IndicIPR-QA test split.
The large gap between BM25 and InLegalBERT shows that lexical matching
alone is insufficient for the benchmark.}
\label{fig:reference_baseline_f1}
\end{figure}

Table~\ref{tab:all_baselines} and Figure~\ref{fig:reference_baseline_f1} summarise the results of all baseline systems using exact match (EM), token level F1, and character level F1 (Char-F1). Standard BM25 retrieval~\cite{robertsonProbabilisticRelevanceFramework2009} achieves a token level F1 of 20.21, while the oracle window setting reaches 51.30. Both remain below the supervised encoder models. Among the encoders,
InLegalBERT achieves the highest F1 (85.02), followed by LegalBERT-US
(84.66) and InCaseLawBERT (84.37). BERT-base has the lowest encoder F1
(83.33). However, the differences between the encoder models are small, so the results do not provide strong evidence that domain specific pretraining alone leads to better performance.

A paired bootstrap significance test with 10,000 resamples and Holm correction finds no statistically significant F1 difference between
InLegalBERT and BERT-base ($\Delta=+1.70$, adjusted $p=0.397$),
LegalBERT-US ($\Delta=+0.36$, adjusted $p=1.000$), or InCaseLawBERT
($\Delta=+0.66$, adjusted $p=1.000$).

Among InLegalBERT's 125 non exact match predictions, 100 (80.0\%) are
span boundary errors, 20 (16.0\%) have low overlap, two (1.6\%) differ
by minor words, two (1.6\%) partially overlap the reference, and one
(0.8\%) is completely wrong. For example, one Copyright prediction
starts with the correct clause from Section~14 but continues into the
next enumerated clause. This shows how long statutory lists can make
it difficult to choose the exact answer boundaries.

\begin{table}[!t]
\centering
\caption{Baseline performance on the IndicIPR-QA test split (344 QA pairs).
ZS = zero-shot; FS = 3-shot few-shot.}
\label{tab:all_baselines}

\scriptsize
\setlength{\tabcolsep}{15pt}
\renewcommand{\arraystretch}{0.9}

\begin{tabular}{@{}lccc@{}}
\toprule

\textbf{System}
& \textbf{EM}
& \textbf{F1}
& \textbf{Char-F1} \\

\midrule

\multicolumn{4}{@{}l}{\textit{Lexical retrieval}} \\
BM25 retrieval     & 0.00  & 20.21 & 77.10 \\
BM25 oracle window & 14.24 & 51.30 & 85.23 \\

\midrule

\multicolumn{4}{@{}l}{\textit{Fine-tuned encoders}} \\
BERT-base          & 63.66 & 83.33 & 86.38 \\
LegalBERT-US       & 65.12 & 84.66 & 86.67 \\
InCaseLawBERT      & 63.08 & 84.37 & 86.40 \\
InLegalBERT        & 63.66 & 85.02 & 87.26 \\

\midrule

\multicolumn{4}{@{}l}{\textit{Prompted Open-Source LLMs}} \\
Mistral 7B (ZS)    & 18.31 & 67.11 & 75.07 \\
Mistral 7B (FS)    & 24.71 & 68.83 & 76.01 \\
Llama 3.1:8B (ZS)  & 33.43 & 74.78 & 79.61 \\
Llama 3.1:8B (FS)  & 33.43 & 77.02 & 79.78 \\
Qwen2.5:7B (ZS)    & 29.94 & 77.59 & 81.04 \\
Qwen2.5:7B (FS)    & 34.30 & 76.39 & 78.60 \\

\bottomrule
\end{tabular}

\end{table}

Among the prompted LLMs, Qwen2.5:7B performs best in the zero-shot
setting (F1 = 77.59), while Llama~3.1:8B performs best in the 3-shot
setting (F1 = 77.02). The best prompted result is still 7.43 F1 points
lower than fine-tuned InLegalBERT, which achieves the highest F1 among
the evaluated models. Section~\ref{sec:generalizability} discusses the scope and
limitations of the generalisation.

\section{Generalizability Analysis}
\label{sec:generalizability}

\begin{figure}[!htbp]
\centering
\includegraphics[width=0.92\linewidth]{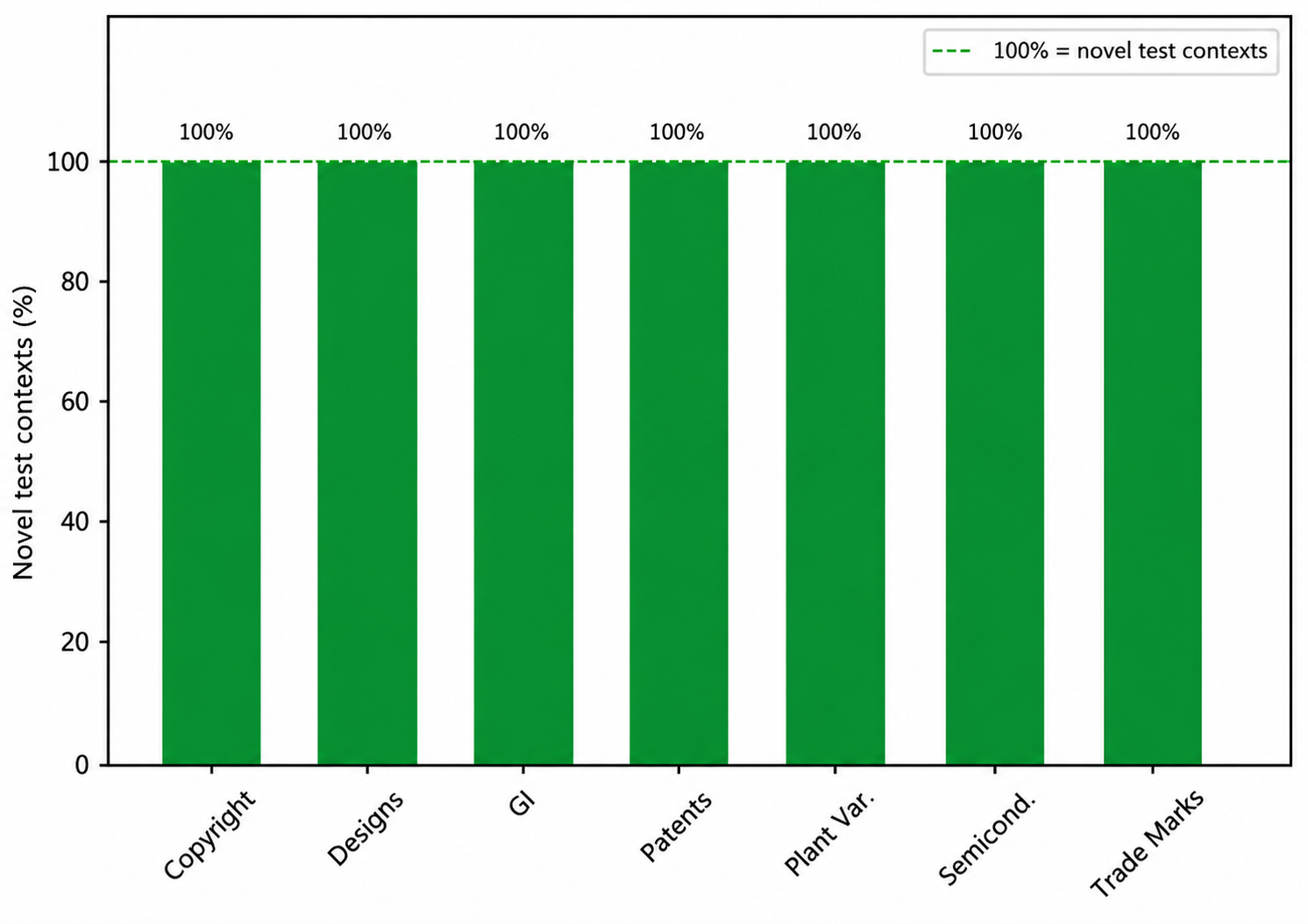}
\caption{Novel test-context rate by Act. The 100\% rate confirms that no source context is shared between the training and test splits.}
\label{fig:generalizability2}
\end{figure}

Section-level splitting prevents the same source context from appearing in both the training and test sets. The train-test context Jaccard similarity is 0.00 for all seven Acts. Figure~\ref{fig:generalizability2} shows that 100\% of test contexts are unseen in the training split for every Act. Among the 1,234 unique answer token types in the test set, 137 (11.1\%) do not appear in the training answers. In addition, 15.3\% of test question tokens and 13.7\% of test context tokens are not found in the training split. This shows that the test set contains unseen legal provisions and words.

However, the questions are not completely independent across the two splits. Of the 4,462 questions, 4,118 are unique after converting them to lowercase, replacing non-word characters with spaces, and removing extra spaces. Common question openings include ``what does'' (1,157), ``under section'' (960), and ``what is'' (875). These repeated GPT-4o question patterns can appear in both training and test sets. Therefore, the results show how well models answer questions generated by the same process over new legal contexts. They do not measure performance on independently written legal questions.

The 64.81-point gap between BM25 and InLegalBERT indicates that simple word matching alone is not sufficient for this task. However, this gap alone does not prove that the models will generalise to real-world legal questions. The BM25 oracle-window F1 score of 51.30 shows that simple lexical matching still performs poorly even when the correct context is provided.




\FloatBarrier

\section{Conclusion and Future Work}
\label{sec:conclusion}

IndicIPR-QA is an extractive QA benchmark containing 4,462 question-answer pairs from 1,116 statutory contexts covering the seven principal Indian IPR Acts. The dataset is created using GPT-4o with nine automatic quality filters, context-balanced selection, model-based evaluation, and a 200-pair human audit. Section-level splitting prevents the same source context from appearing in both the training and test sets. Using the unified sliding-window inference method, fine-tuned InLegalBERT achieves a token-level F1 of 85.02, while BM25 retrieval achieves 20.21. This large difference shows that simple word matching is not enough for this task. Performance also differs across the seven Acts and different types of answers, and most InLegalBERT errors are related to choosing the exact answer boundaries. These results highlight the difficulty of precise answer-span extraction, especially for long and multi-clause answers. IndicIPR-QA provides a benchmark for studying extractive QA over Indian IPR law. The benchmark is limited to English, answerable, single-context questions, and the questions are generated using GPT-4o. Future work can include independently written questions, more Indian languages,
unanswerable questions, and scenario based or multi section questions.

\section{Data and Code Availability}

The IndicIPR-QA dataset and code are publicly available:

\textbf{Dataset:} \url{https://huggingface.co/datasets/niloycste68/IndicIPRExtractiveQA}

\textbf{Code:} \url{https://github.com/niloycste/Indic-IPR-Dataset-}


\balance
\bibliographystyle{IEEEtran}
\bibliography{dsaa_references}

\end{document}